\documentclass{article}
\usepackage{ijcai24}

\usepackage{times}
\usepackage{soul}
\usepackage{url}
\usepackage[hidelinks]{hyperref}
\usepackage[utf8]{inputenc}
\usepackage[small]{caption}
\usepackage{graphicx}
\usepackage{amsmath}
\usepackage{amsthm}
\usepackage{booktabs}
\usepackage{algorithm}
\usepackage{algorithmic}
\usepackage[switch]{lineno}
\usepackage{color}

\usepackage{threeparttable}
\usepackage{booktabs}
\usepackage{threeparttable}
\usepackage{multicol}
\usepackage{multirow}
\usepackage{amsmath}
\usepackage{amssymb}
\usepackage{graphicx}
\usepackage{caption,subcaption}
\usepackage{makecell}
\usepackage{colortbl}
\usepackage{xcolor}

\title{Toward Robust Multimodal Learning using\\
Multimodal Foundational Models}

\author{
Xianbing Zhao$^1$
\and
Soujanya Poria$^2$\and
Xuejiao Li$^{1}$\And
Yixin Chen$^1$\\
Buzhou Tang$^1$\\
\affiliations
$^1$Harbin Institute of Technology Shenzhen\\
$^2$Singapore University of Technology and Design, Singapore\\
\emails
zhaoxianbing\_hitsz@163.com,
sporia@sutd.edu.sg,
lixj269@mail2.sysu.edu.cn,
tangbuzhou@gmail.com
}

\begin{document}

\maketitle

\begin{abstract}
Existing multimodal sentiment analysis tasks are highly rely on the assumption that the training and test sets are complete multimodal data, while this assumption can be difficult to hold: the multimodal data are often incomplete in real-world scenarios. Therefore, a robust multimodal model in scenarios with randomly missing modalities is highly preferred. Recently, CLIP-based multimodal foundational models have demonstrated impressive performance on numerous multimodal tasks by learning the aligned cross-modal semantics of image and text pairs, but the multimodal foundational models are also unable to directly address scenarios involving modality absence. To alleviate this issue, we propose a simple and effective framework, namely TRML, Toward Robust Multimodal Learning using Multimodal Foundational Models.  TRML employs generated virtual modalities to replace missing modalities, and aligns the semantic spaces between the generated and missing modalities. Concretely, we design a missing modality inference module to generate virtual modaliites and replace missing modalities. We also design a semantic matching learning module to align semantic spaces generated and missing modalities.  Under the prompt of complete modality, our model captures the semantics of missing modalities by leveraging the aligned cross-modal semantic space. Experiments demonstrate the superiority of our approach on three multimodal sentiment analysis benchmark datasets, CMU-MOSI, CMU-MOSEI, and MELD.

\end{abstract}
\maketitle

\section{Introduction}
Human often perceive the real world in daily life through multimodal signals, such as language, visual, and acoustic signals,\begin{figure}[t]
  \centering
  \includegraphics[width=\linewidth]{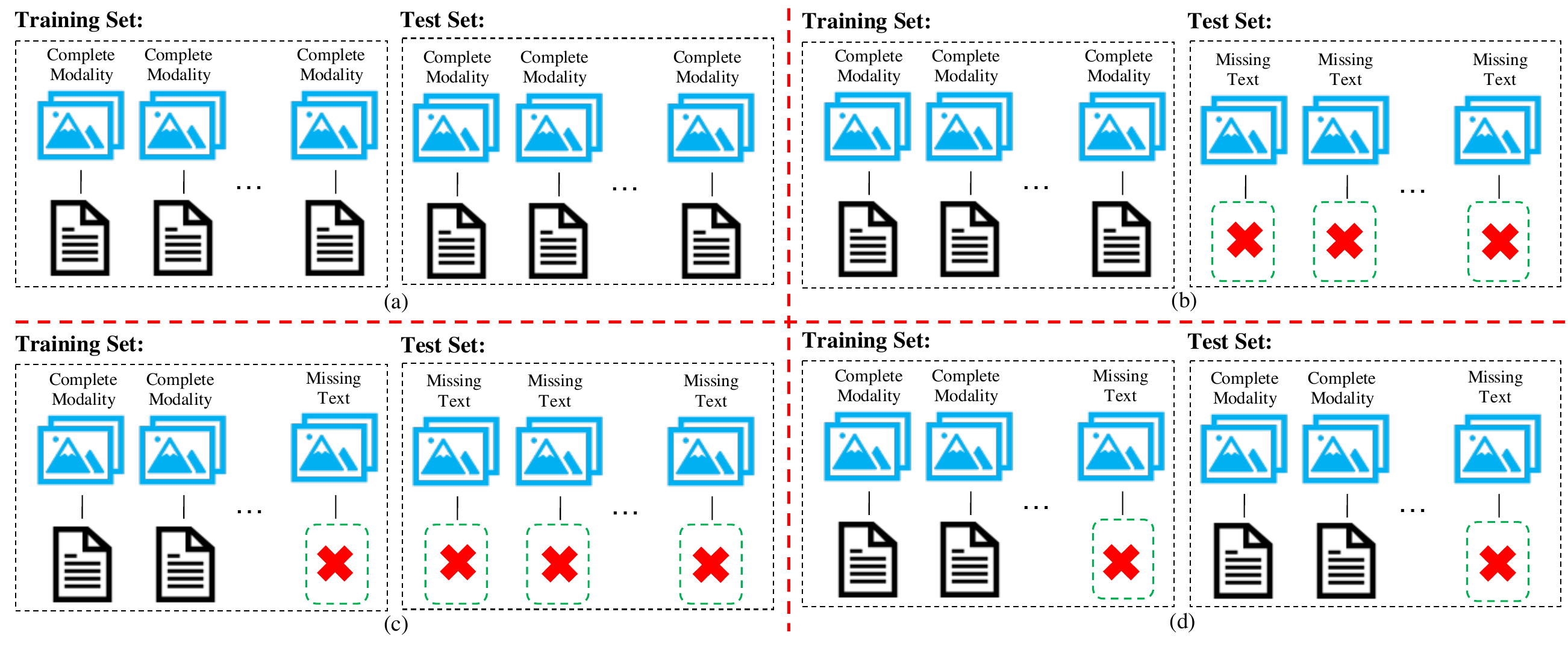}
  \caption{Scenarios with missing modalities. Taking the missing text modality as an example, black indicates that the visual modality is the victim modality. (a): Modalities are complete. (b): Modalities are complete in the training set and victim modality is completely missing in the test set. (c): Modalities are missing in both training and test sets. Victim modality is missing randomly in the training set and completely missing in the test set. (d)  Modalities are missing randomly in both training and test sets with the same probability.} 
  \label{fig:premodel}
\end{figure}
 thus, modeling and cooperating with multiple types of signals in multimodal learning is a topic that has attracted much attention and has great application potential \cite{zadeh2016mosi,zadeh2017tensor,zadeh2018multimodal,tsai2019multimodal,rahman2020integrating,yu2021learning,zhao2022shared,li2023blip,zhang2023prompt}.  Although considerable performance has been achieved in the field of multimodal learning, most of these multimodal learning models consider the perfect application scenario: complete multimodal input settings. In the practical scenario, the modality missing setting may occur in either the training set or the test set, that is, the incomplete training data and test data \cite{pham2019found,zhao2021missing,ma2022multimodal,lee2023multimodal}. 

Based on the assumption of complete modality input (i.e., Fig. \ref{fig:premodel}a), existing works mainly focuses on designing complex multimodal fusion methods \cite{zadeh2017tensor,zadeh2018multimodal,tsai2019multimodal,rahman2020integrating,zhao2022mag+} and multimodal representation learning \cite{hazarika2020misa,yu2021learning,yang2022disentangled} methods for training and inference in the scenario of complete data. Based on the common assumption of data incompleteness, existing works mainly use auxiliary tasks to calculate cross-modal information transfer to reconstruct missing modalities \cite{pham2019found,wang2020transmodality,ma2021smil,han2022mm}. However, there
are still challenges for missing modality inference of multimodal learning being applied in real-world scenarios: cross-modal information transfer is difficult due to misaligned semantic spaces across modalities. Recently, many multimodal foundational models using large-scale image-text pairs have shown outstanding zero-shot ability for various multimodal tasks by learning a jonint embedding spaces of image-text pairs \cite{radford2021learning,zheng2022general,ni2022expanding,li2023blip,girdhar2023imagebind}. 
This indicates that the multimodal foundational models are naturally robust for cross-domain tasks because they learn the aligned the semantic spaces across modalities. However, their limitations lie in the lack of capacity to handle missing modality during training and testing phases. Recent works on robust multimodal learning using multimodal foundational models is seldom investigated in the literature. The latent semantic correlations provided by multimodal foundational models are crucial for leveraging complete modalities to generate missing modalities.
 
To mitigate this issue, we propose the TRML , a novel and simple towards robust multimodal learning framework, which extends the of multimodal foundational models to address scenarios involving missing modality. As shown in Fig. 2, the TRML  consists of three components. 1) We embed different multimodal foundational models in our framework to learn latent semantic correlated multimodal representations. 2) We tailor the missing modality inference module to generate the \textbf{virtual visual/text modality}  and replace the victim modality in the scenario of missing visual/text modality. 3) We design a semantic matching learning module to align the semantic spaces of virtual visual/text and remain visual/text.  Constrained by the semantic matching module, the model-generated virtual visual/text modalities can learn the semantic information of missing visual/text modalities under the prompt of complete text/visual modality. The virtual modalities are utilized in training and testing when victim modalities are missing.
Fig. \ref{fig:premodel}(c) and Fig \ref{fig:premodel}(d) display the scenario if missing modality. In scenarios with incomplete modalities, we use virtual modalities to replace the missing modalities. Extensive experimental results on three multimodal sentiment analysis benchmark datasets, i.e., CMU-MOSEI, CMU-MOSI, and MELD, validate the demonstrate the robustness and effectiveness of our proposed framework.

The main contributions of this work are three-fold:
\begin{itemize}
    \item We design a simple yet robust  multimodal learning framework. The framework extends the capabilities of the multimodal foundational models to address the scenarios involving missing modalities by utilizing the latent semantic association, thereby enhancing robustness.
    \item We design missing modality inference module to generate virtual visual/text modality, and tailor a semantic matching learning module to align the semantic spaces of virtual visual/text and remain visual/text modality, which ensures that the generated virtual modalities can learn the semantics of missing modalities.
    \item Our model achieves a more prominent performance than previous works. In addition, compared with the state-of-the-art work, our model can deal with scenes that are missing at the frame level, which is also the limitation of MM-align.
    
\end{itemize}

\section{Related Work}
\subsection{Multimodal Learning}
Multimodal learning utilizes complementary multimodal data to learn a common concept as biological systems perceive multimodal signals. Existing multimodal learning methods are mainly divided into two categories: multimodal fusion method and multimodal representation learning method. Multimodal fusion methods attempt to design complex multimodal interaction patterns to complete the information exchange across modalities. Existing multimodal fusion methods span from designing tensor fusion network \cite{zadeh2017tensor,liu2018efficient} to cross-attention interaction patterns \cite{zadeh2018memory,zadeh2018multimodal,tsai2019multimodal,rahman2020integrating,lv2021progressive,yang2021mtag,han2021improving,zhao2022mag+} to capture task-related information. Multi-modal representation learning methods attempt to learn modal-specific and modal-common representations for the characteristics of multimodal data  by designing different auxiliary tasks \cite{hazarika2020misa,yang2022disentangled,zhao2022shared}. The premise of both multimodal representation learning methods and multimodal fusion methods is complete multimodal data input. There is a small body of literature that attempts to explore scenarios with missing modalities. They mainly focus on employing an auxiliary task of cross-modal information transfer to reconstruct missing modality \cite{pham2019found,wang2020transmodality,ma2021smil,han2022mm}. These methods primarily generate missing modalities in the unaligned semantic space. Distinguishing from them, our proposed model ensures that missing modalities learn the semantics of the original modality by aligning cross-modal semantic spaces.

\subsection{Multimodal Foundational Models}

Multimodal foundational models achieved considerable performance \cite{miech2019howto100m,sun2019learning,zhu2020actbert,radford2021learning,ni2022expanding,li2023blip,girdhar2023imagebind}, which aligns and binds multimodal representations in a joint embedding space. They benefit from large-scale contrastive language-image pretraining being able to learn cross-modal semantic associations. The representative work is CLIP \cite{radford2021learning}, which has shown promise ability for multimodal tasks \cite{frome2013devise,kiros2014unifying,socher2014grounded,faghri2017vse++}. The excellent performance of CLIP has inspired approaches to learning deep semantic associations across modalities. Other achievements transfer this technique to the video domain to learn deep semantic matching information between video frames and language \cite{zheng2022general,ni2022expanding} and richer multimodal signals to bind multimodal representation in a joint embedding space. While the multimodal foundational model has achieved impressive performance, it exhibits limitations in handling missing modality. Our goal is to address the limitations of the multimodal foundational models.

\begin{figure*}
    \centering
    \includegraphics[width=\linewidth]{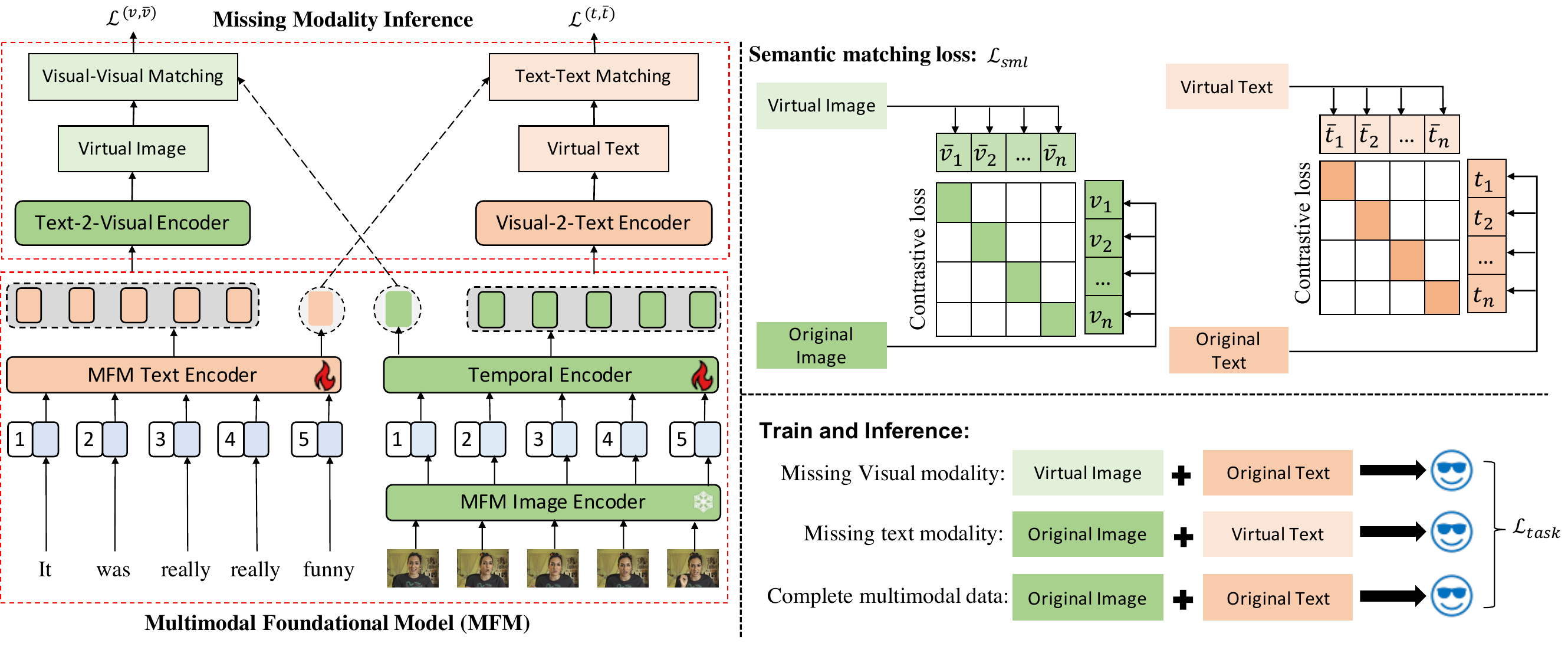}
    \caption{ Schematic illustration of the proposed TRML  framework. Our model comprises three components: 1) Multimodal Foundational Model that learns representations for latent semantic alignment; 2) Missing Modality Inference module, taking missing text modality as an example, utilizing visual modality as prompt to generate a virtual missing text modality; 3) Semantic Match Learning module aligns the semantic space of the virtual text modality with the original text modality, enabling virtual modality to learn the similar semantic aspects of the missing text modality.
    }
    \label{fig:model}
\end{figure*}
\section{Method}
\subsection{Problem Definition}
In the scenario of complete modalities, the goal of multimodal learning method is to learn the mapping relationship from $\mathcal{X}$ $\rightarrow$ $\mathcal{Y}$ by allowing the input of multiple modalities $x^{m_1}$ and $x^{m_2}$. $m_1$ and $m_2$ represent the modal category, such as text and visual modality. $\mathcal{X}$ represents multimodal space, and $\mathcal{Y}$ denotes label space. The multimodal dataset is divided into training data, validation data, and test data $\{\mathcal{D}^{train}, \mathcal{D}^{valid}, \mathcal{D}^{test}\}$. $\mathcal{D}^{train}=\{(x_{j}^{m_1},x_{j}^{m_2};y_{j})_{j=1}^{n_1}\}$ and $\mathcal{D}^{test}=\{(x_{k}^{m_1},x_{k}^{m_2};y_{})_{k=1}^{n_2}\}$ denotes the samples and labels of training data and test data, respectively. $n_1$ and $n_2$ denote the number of samples for training and testing data, respectively. In the scenario of incomplete modalities,  we consider multimodal datasets with varying degrees of missing.
Following previous works \cite{ma2021smil,han2022mm}, we refer to the \textbf{missing modality} as the \textbf{victim modality}. The victim modality may be present in the training set or the test set. Consistent with the previous work, we consider two different modalities missing settings: 1) '\textbf{Setting A}', the victim modality is completely missing in the test set (corresponding to Fig. \ref{fig:premodel}(c)). 2) '\textbf{Setting B}', the victim modality is randomly missing (corresponding to Figure \ref{fig:premodel}(d)) in the training set and the test set with the same probability. \textbf{Mark $A\rightarrow B$ denotes that victim modality is modality B, and we use complete modality A to infer victim modality B}.

\subsection{Model Overview}
Our goal is to enhance the capabilities of the multimodal foundational model in both training and testing scenarios with missing modality. Our model contains three components: 1) \textbf{Multimodal Foundational Model (MFM)}. The multimodal foundational model learns cross-modal semantic associations during contrastive pre-training. Thus, we exploit the encoder of multimodal foundational model to learn latent semantic association of image and text paires. 2) \textbf{Missing Modality Inference}. We tailor a Text-2-Visual Encoder and a Visual-2-Text Encoder to generate virtual visual and text modality. When scenarios with missing modalities are presented, we use virtual modalities to replace the missing modalities. and 3) \textbf{Semantic Matching Learning}. To enable the missing modality to learn the semantics of the original modality, we  design a semantic matching module to align the cross-modal semantic spaces of the missing modality and the original modality under the prompt of complete modalities. 

\subsection{Missing Modality Inference}
Our model employs the multimodal foundational models \cite{radford2021learning,ni2022expanding,girdhar2023imagebind} for missing modality inference. To unify the input form, we only utilize the text encoder and the visual encoder of the multimodal foundational model to learn latent semantic associated features (i.e., $x_t$ and $x_{\Bar{v}}$) of the input text modality (T) and video frames (V) with parameters $\theta_{t}$ and $\theta_{\bar{v}}$.
\begin{gather}
    x_{t} = f_{\theta_t}(T)\\
    x_{\bar{v}} = f_{\theta_{\bar{v}}}(V)
\end{gather}
where $x_t\in \mathbb{R}^{1\times d}$ and $x_{\bar{v}}\in \mathbb{R}^{n\times d}$ represents the features of text modality and video frames. The superscripts $n$ and $d$ denote the number of video frames and feature dimension, respectively. To capture the temporal correlation of video frames, we adopt a temporal encoder to process the representation of video frames with parameters $\theta_{v}$ and obtain the representation $x_{v}$ of visual modality. Formally,
\begin{equation}
    x_{v} = f_{\theta_v}(x_{v}), x_{v}\in\mathbb{R}^{1\times d},
\end{equation}
where the Temporal Encoder consists of an RNN encoder.  To perform missing modality inference, we tailor two encoders ($f_{\theta_{v2t}}$,$f_{\theta_{t2v}}$) to generate virtual modalities ($\Bar{x}_t$,$\bar{x}_v$) under the prompt of the corresponding original modality ($x_v$,$x_t$). The virtual modalities are then utilized in training and testing, replacing the original modality when modalities are missing. Formally,
\begin{gather}
    \bar{x}_{t}=f_{\theta_{v2t}}(x_{t})\\
    \bar{x}_{v}=f_{\theta_{t2v}}(x_{v})\\
     f_{\theta_{v2t/t2v}} = \mathrm{MLP}(\mathrm{ReLU}(\mathrm{MLP}(x_{t/v})))
\end{gather}
when the original modality transforms into an afflicted modality, the virtual modality is engaged in both training and testing. Based on the missing modality inference module, we enhance the robustness of multimodal foundational model with scenarios of missing modality during training and test phrase. Formally,
\begin{gather}
    x = x_{t} + \bar{x}_{v}, mv;\\
    x = x_{t} + \bar{x}_{v}, mt;\\
    x = x_{t} + x_v, c;
\end{gather}
where $mv$ and $mt$ represent missing visual modality and text modality, respectively. $c$ denotes complete multimodal data. $x$ is the final representation of multimodal data, and it passes through several fully connected layers for multimodal sentiment analysis and emotion recognition tasks.

\subsection{Semantic Matching Learning}
Previous work\cite{radford2021learning} has shown that contrastive learning has a powerful ability to align deep semantic associations across modalities. Motivated by this, we tailor a semantic matching learning module to the deep semantic information between original modalities and virtual modalities. To align the semantics of the virtual modality with the original modality (i.e., ($x_t,\bar{x}_t$) and ($x_v,\bar{x}_v$,)), the semantic matching learning module aligns the cross-modal semantic space under the prompt of original modality, fully leveraging the potential cross-modal semantic correlations inherent in the multimodal foundational model.  In the training phase, we calculate the similarity matrix $s^{t}$ of the original text modality $x_t$ and the virtual text modality $\bar{x}_t$ for Text-Text Matching. Similarity, we also calculate the similarity matrix $s^{v}$ of the original visual modality $x_v$ and the virtual visual modality $\bar{x}_v$ for Visual-Visual Matching. Formally,
\begin{gather}
    s^{t}=<x_{t},\bar{x}_{t}>=\frac{x_{t}^{T}\cdot \bar{x}_{t}}{||x_{t}||\cdot ||\bar{x}_{t}||}\\
    s^{v}=<x_{v},\bar{x}_{v}>=\frac{x_{v}^{T}\cdot \bar{x}_{v}}{||x_{v}||\cdot ||\bar{x}_{v}||}
\end{gather}
where the similarity matrix $s^{t}$ and $s^{v}$ represent the Text-Text Matching and Visual-Visual Matching Scores.

In a mini-batch N, for any original text modality $x_{i}$, $s^{t}$ is the similarity between the original text modality $x_{i}$ and the virtual text modality $x_{j}$. The subscript $i$ and $j$ denote the i-th and j-th samples in the mini-batch. Therefore, we obtain the normalized semantic similarity matrix of the rows (i.e.,$t\rightarrow \bar{t}$) and columns (i.e.,$\bar{t}\rightarrow t$) by computing softmax. Formally,
\begin{gather}
    y_{ij}^{t\rightarrow \bar{t}} = \frac{exp(s^{t}_{ij}/\tau)}{\sum_{j=1}^{N}exp(s_{ij}^{t}/\tau)}\\
    y_{ji}^{\bar{t}\rightarrow t} = \frac{exp(s^{t}_{ji}/\tau)}{\sum_{i=1}^{N}exp(s_{ji}^{t}/\tau)}
\end{gather}
where $\tau$ is a learnable temperature parameter. Similarly, we also normalize the semantic similarity matrix $S^{v}$ of the original visual modality  and the virtual visual modality by computing the softmax. Formally,

\begin{gather}
    y_{ij}^{v\rightarrow \bar{v}} = \frac{exp(s^{v}_{ij}/\tau)}{\sum_{j=1}^{N}exp(s_{ij}^{v}/\tau)}\\
    y_{ji}^{\bar{v}\rightarrow v} = \frac{exp(s^{v}_{ji}/\tau)}{\sum_{i=1}^{N}exp(s_{ji}^{v}/\tau)}
\end{gather}
where $v\rightarrow \bar{v}$ and $\bar{v}\rightarrow v$ denote the normalization of the rows and columns of the similarity matrix, respectively.
Let $\hat{y}^{t\rightarrow \bar{t}}$, $\hat{y}^{\bar{t}\rightarrow t}$, $\hat{y}^{v\rightarrow \bar{v}}$, and $\hat{y}^{\bar{v}\rightarrow v}$ denote the ground-truth one-hot similarity. The original modality and virtual modality (i.e. $(t,\bar{t})$ and $(v,\bar{v})$) of the same sample are regarded as positive samples, and the original modality and virtual of different samples are regarded as negative samples. The probability value of positive samples tends to be $1$, and the probability value of negative samples tends to be $0$. The semantic matching loss can be achieved by cross-entropy loss. Formally,
\begin{minipage}{\linewidth}
\begin{gather}
\begin{split}
    \mathcal{L}^{(t,\bar{t})}=\frac{1}{2}\mathbb{E}_{(t,\bar{t})\sim D}[    H(y^{t\rightarrow\bar{t}},\hat{y}^{t\rightarrow\bar{t}})\\ + H(y^{\bar{t}\rightarrow t},\hat{y}^{\bar{t}\rightarrow t})]
    \end{split}\\
    \begin{split}
    \mathcal{L}^{(v,\bar{v})}=\frac{1}{2}\mathbb{E}_{(v,\bar{v})\sim D}[    H(y^{v\rightarrow\bar{v}},\hat{y}^{v\rightarrow\bar{v}})\\ + H(y^{\bar{v}\rightarrow v},\hat{y}^{\bar{v}\rightarrow v})]
    \end{split}
\end{gather}
\end{minipage}
where $H$ denotes cross-entropy loss. In the training phase, the final training objective of semantic matching loss is to integrate these two losses using weights. Formally,
\begin{equation}
    \mathcal{L}_{sml} = \lambda \mathcal{L}^{(t,\bar{t})} + (1-\lambda) \mathcal{L}^{(v,\bar{v})}
\end{equation}
where $\lambda$ is an adjustable hyper-parameter that takes the range (0,1). We use mean absolute error (MSE) and cross entropy as the loss function $\mathcal{L}_{task}$ on multimodal sentiment analysis and multimodal emotion recognition tasks, respectively. The objective function is as the following form:
\begin{equation}
    \mathcal{L} = \mathcal{L}_{task} + \alpha \cdot \mathcal{L}_{sml}
\end{equation}
where $\alpha$ is a hyper-parameter.

\section{Experiments}
The detail of \textbf{Dataset}, \textbf{Baselines}, and \textbf{Experimental settings} can be found in the \textbf{Appendix} of supplementary materials. Regarding the missing data setting, we consider the \textbf{most extreme missing modality scenario}, where missing modalities occur in both the training and test sets (that is, Fig. \ref{fig:premodel}(c) and Fig. \ref{fig:premodel}(d) ). Following previous works \cite{ma2021smil,han2022mm}, we also set a probability $p^{'}=1-p$ to indicate the proportion of randomly removed data (the missing data of victim modality). The victim modality for \textbf{Setting A} (i.e., Fig. \ref{fig:premodel}c) are partially missing for the training set and completely missing for the test set. The victim modality for \textbf{Setting B} is partially missing for the training set and completely missing for the test set.

\begin{table}
    \centering
    \scriptsize
    \setlength{\tabcolsep}{4mm}
    \begin{tabular}{ccccccccc}
    \toprule 
    \multirow{3}{*}{ Method } & \multicolumn{4}{c}{$\mathrm{T} \rightarrow \mathrm{V}$ (Missing visual modality.)}  \\
    \cmidrule(r){2-5} 
    & \multicolumn{2}{c}{ Setting A } & \multicolumn{2}{c}{ Setting B } \\
    \cmidrule(r){2-3}  \cmidrule(r){4-5} 
    & MAE $\downarrow$ & Acc-2 $\uparrow$ & MAE $\downarrow$ & Acc-2 $\uparrow$ \\
    \midrule
    LB  & 0.802 & 81.5 & 0.802 & 81.5  \\
    UB  & 0.751 & 86.36 & 0.751 & 86.36 \\
    \midrule
    MFM & 1.103 & 71.0 & 1.093 & 73.2  \\
    SMIL & 1.073 & 74.2 & 1.052 & 75.3  \\
    Modal-Trans & 1.052 & 75.5 & 1.041 & 75.8 \\
     MM-Align & 1.028 & 76.9 & 1.027 & 77.0  \\
    \midrule
    \textbf{TRML}   & \textbf{0.757} & \textbf{86.0} & \textbf{0.755} & \textbf{86.3}  \\
    \bottomrule
    \end{tabular}
    \caption{Performance comparison on CMU-MOSI dataset. The value of $p$ is set to 10\% (i.e., the proportion of missing modality is 90\%).     
    }
    \label{tab:mosi-t}
\end{table}

\begin{table}
    \centering
    \scriptsize
    \setlength{\tabcolsep}{4mm}
    \begin{tabular}{ccccccccc}
    \toprule 
    \multirow{3}{*}{ Method } & \multicolumn{4}{c}{$\mathrm{V} \rightarrow \mathrm{T}$ (Missing text modality.)}  \\
    \cmidrule(r){2-5} 
    & \multicolumn{2}{c}{ Setting A } & \multicolumn{2}{c}{ Setting B }  \\
    \cmidrule(r){2-3}  \cmidrule(r){4-5} 
    & MAE $\downarrow$\ & Acc-2 $\uparrow$ & MAE $\downarrow$ & Acc-2 $\uparrow$ \\
    \midrule
    LB   & 1.47 & 53.2 & 1.47 & 53.2 \\
    UB  & 0.751 & 86.36 & 0.751 & 86.36 \\
    \midrule
    MFM & 1.479 &42.2 &1.429 &51.9 \\
    SMIL  &1.448 &44.2 &1.447 &43.3 \\
    Modal-Trans  &1.429 & 50.3 &1.420 &53.1 \\
     MM-Align  & 1.415 & 52.7& 1.410 & 53.4 \\
    \midrule
    \textbf{TRML}   & \textbf{1.410} & \textbf{61.5} & \textbf{1.408} & \textbf{61.9} \\
    \bottomrule
    \end{tabular}
    \caption{Performance comparison on CMU-MOSI dataset. The value of $p$ is set to 10\% (i.e., the proportion of missing modality is 90\%). 'C' denotes that the multimodal foundational model is CLIP.}
    \label{tab:mosi-v}
\end{table}

\begin{table}
    \centering
    \scriptsize 
    \setlength{\tabcolsep}{4mm}
    \begin{tabular}{ccccccccc}
    \toprule 
    \multirow{3}{*}{ Method } & \multicolumn{4}{c}{$\mathrm{T} \rightarrow \mathrm{V}$ (Missing visual modality.)} \\
    \cmidrule(r){2-5}
    & \multicolumn{2}{c}{ Setting A } & \multicolumn{2}{c}{ Setting B }\\
    \cmidrule(r){2-3} \cmidrule(r){4-5}
    & MAE $\downarrow$ & Acc-2 $\uparrow$ & MAE $\downarrow$ & Acc-2 $\uparrow$  \\
    \midrule
     LB & 0.603 & 83.8 & 0.603 & 83.8 \\
    UB & 0.581 & 85.3 &  0.581 & 85.3 \\
    \midrule
    MFM & 0.658 & 79.2 &0.645 &80.0  \\
    SMIL & 0.680 &78.3 &0.648 &78.5 \\
    Modal-Trans & 0.645 & 79.6 &0.647 &79.6  \\
    MM-Align & 0.637 & 80.8 & 0.638 & 81.1   \\
    \midrule
    \textbf{TRML}   &   \textbf{0.590} & \textbf{84.8} & \textbf{0.582} & \textbf{85.1} \\
    \bottomrule
    \end{tabular}
    \caption{The performance comparison on CMU-MOSEI ($p$=10\%). Notations share the same meaning as the last table. 
    \label{tab:mosei-t}
    }
\end{table}

\begin{table}
    \centering
    \scriptsize
    \setlength{\tabcolsep}{4mm}
    \begin{tabular}{ccccc}
    \toprule
    \multirow{3}{*}{ Method } & \multicolumn{4}{c}{$\mathrm{V} \rightarrow \mathrm{T}$ (Mssing text modality.)} \\
    \cmidrule(r){2-5} 
    & \multicolumn{2}{c}{ Setting A } & \multicolumn{2}{c}{ Setting B }  \\
    \cmidrule(r){2-3} \cmidrule(r){4-5}
    &  MAE $\downarrow$\ & Acc-2 $\uparrow$ & MAE $\downarrow$ & Acc-2 $\uparrow$ \\
    \midrule
    LB & 0.874 & 72.7 & 0.874 & 72.7 \\
    UB  &  0.581 & 85.3 &  0.581 & 85.3 \\
    \midrule
    MFM & 0.821 &62.0 &0.817 &61.7 \\
    SMIL  & 0.820& 63.1 &0.816 &63.5 \\
    Modal-Trans & 0.817 & 65.1 &0.814 &65.7 \\
     MM-Align  & 0.811 & 66.2 & 0.806 & 66.9  \\
    \midrule
    \textbf{TRML}    & \textbf{0.856} & \textbf{73.9} & \textbf{0.818} & \textbf{75.8} \\
    \bottomrule
    \end{tabular}
    \caption{The performance comparison on CMU-MOSEI ($p$=10\%).
    \label{tab:mosei-v}
    }
\end{table}

\begin{table}[]
    \centering
    \scriptsize
    \setlength{\tabcolsep}{0.8mm}
    \begin{tabular}{ccccc}
    \toprule 
    \multirow{3}{*}{ Method }& \multicolumn{2}{c}{$\mathrm{T} \rightarrow \mathrm{V}$ (Missing visual modality)} & \multicolumn{2}{c}{$\mathrm{V} \rightarrow \mathrm{T}$ (Missing text modality)}   \\
    \cmidrule(r) { 2 - 3 }  \cmidrule(r) { 4 - 5 }
    &Setting A & Setting B & Setting A & Setting B \\
    \cmidrule(r){2-3} \cmidrule(r){4-5}
     & Acc-2 $\uparrow$ & Acc-2 $\uparrow$ & Acc-2 $\uparrow$ & Acc-2 $\uparrow$ \\
    \midrule 
    LB & 57.4 & 57.4 & 42.2 & 42.2 \\
    UB & 64.9 & 64.9 & 64.9 & 64.9 \\
    \midrule
    MFM & 54.0 & 53.9 & 31.4 & 43.6 \\
    SMIL & 54.4 & 54.2 & 31.4 & 43.9 \\
    Modal-Trans & 55.0 & 54.8 & 31.6 & 44.2 \\
    MM-Align & 55.7 & 55.7 & 32.3& 45.4\\
    \midrule
    \textbf{TRML}   & \textbf{63.5} & \textbf{64.2} & \textbf{46.1} & \textbf{46.2}\\
    \toprule
    \end{tabular}
    \caption{The performance comparison on MELD.
    }
    \label{tab:meld}
\end{table}

\subsection{Overall Results}
To verify the effectiveness of our model in missing modal inference, we compare several baselines for missing modal inference. The performance of our model on the three multimodal benchmark datasets (CMU-MOSI, CMU-MOSEI, and MELD) are summarized in Table \ref{tab:mosi-t}, \ref{tab:mosi-v}, \ref{tab:mosei-t}, \ref{tab:mosei-v} and \ref{tab:meld}. We adopt CLIP as the multimodal foundational model. By analyzing the above tables, we can draw the following conclusions:
\begin{itemize}
    \item Our model achieves an improvement over the previous method on three multimodal benchmark datasets.
    \item Our model demonstrates performance closer to the upper bound when visual modality is missing. There is a performance improvement of 1\%$\sim$11\% when the text modality is missing relative to the lower bound.
    \item Our model approaches the upper bound of full modality when the visual modality is missing. In both Setting A and Setting B, our model exhibits greater stability and demonstrates stronger robustness compared to other models in terms of performance.
\end{itemize}
This fact indicates that elaborately establishing missing modality inference  and semantic matching learning modules based on multimodal foundational models is extremely essential.  Additionally, we incorporate the latest multimodal foundational model to include the audio modality in our training. Figure \ref{fig:imagebind} illustrates the performance of our framework on Imagebind. This result further highlights the scalability of our framework.

\begin{table}[]
    \centering
    \tiny
    \begin{tabular}{cccccccccc}
    \toprule
    \multirow{3}{*}{Variants} & \multirow{3}{*}{Dataset} & \multicolumn{4}{c}{ T$\rightarrow$V (Missing visual modality)}\\
    \cmidrule(r){3-6}
    & & \multicolumn{2}{c}{Setting A} & \multicolumn{2}{c}{Setting B}  \\
    \cmidrule(r){3-4} \cmidrule(r){5-6}
    & & MAE $\downarrow$ & Acc-2 $\uparrow$ & MAE $\downarrow$ & Acc-2 $\uparrow$  \\
    \midrule
    \textbf{TRML}  & \textbf{CMU-MOSI} & \textbf{0.757} & \textbf{86.0} & \textbf{0.755} & \textbf{86.3}  \\
    $w / o \mathcal{L}^{(t, \bar{t})}$ & CMU-MOSI & 0.777 & 85.6 & 0.763 & 85.9  \\
    $w / o \mathcal{L}^{(v, \bar{v})}$ & CMU-MOSI & 0.791 & 84.8 & 0.789 & 85.3  \\
    $w / o \mathcal{L}_{s m l}$ & CMU-MOSI & 0.793 & 84.5 & 0.792 & 84.7 \\
    \midrule
    \textbf{TRML}  & \textbf{CMU-MOSEI} & \textbf{0.590} & \textbf{84.8} & \textbf{0.582} & \textbf{85.1}  \\
    $w / o \mathcal{L}^{(t, \bar{t})}$ & CMU-MOSEI & 0.593 & 84.3 & 0.595 & 84.4  \\
    $w / o \mathcal{L}^{(v, \bar{v})}$ & CMU-MOSEI & 0.594 & 84.5 & 0.588 & 84.6  \\
    $w / o \mathcal{L}_{s m l}$ & CMU-MOSEI & 0.597 & 83.9 & 0.595 & 84.2  \\
    \bottomrule
    \end{tabular}
    \caption{Ablation experiments for missing visual modality $V$.}
    \label{tab:ablation-t}
\end{table}

\subsection{Ablation Study}
To gain insight into the capabilities of our multimodal learning framework TRML , we conducted ablation experiments from two aspects. First, we adopt different multimodal foundational models under the framework of TRML , which include CLIP~\cite{radford2021learning}, X-CLIP~\cite{ni2022expanding}, and ImageBind~\cite{girdhar2023imagebind}. Second, we test the ability of our designed semantic matching loss to  motivate the robustness of  the multimodal foundational model. We compared our model with the following variants: 
1) \textbf{w/o $\mathcal{L}^{t,\bar{t}}$}, removing the semantic similarity loss of original text modality and virtual text modality. 
2) \textbf{w/o $\mathcal{L}^{v,\bar{v}}$}, semantic similarity loss of original visual modality and virtual visual modality  were removed. 
3) \textbf{w/o $\mathcal{L}_{sml}$}, excluding semantic matching loss retaining the encoder of virtual.
\begin{figure}
    \centering
    \includegraphics[width=\linewidth]{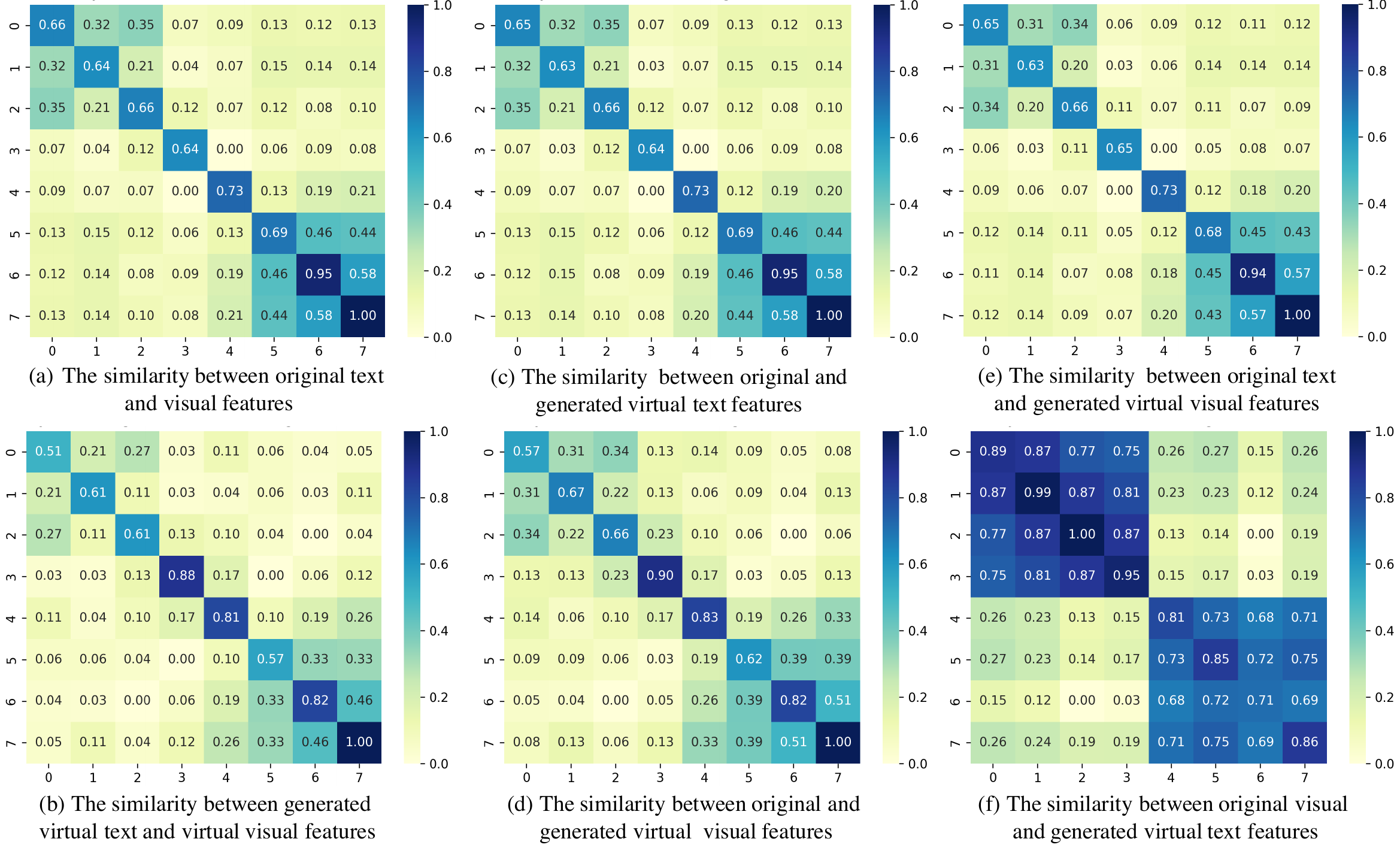}
    \caption{The semantic similarity matrix between the generated virtual modality and the virtual modality of randomly selected 8 samples on the CMU-MOSI dataset.}
    \label{fig:similarity}
\end{figure}

\begin{figure}
    \centering
    \includegraphics[width=6cm]{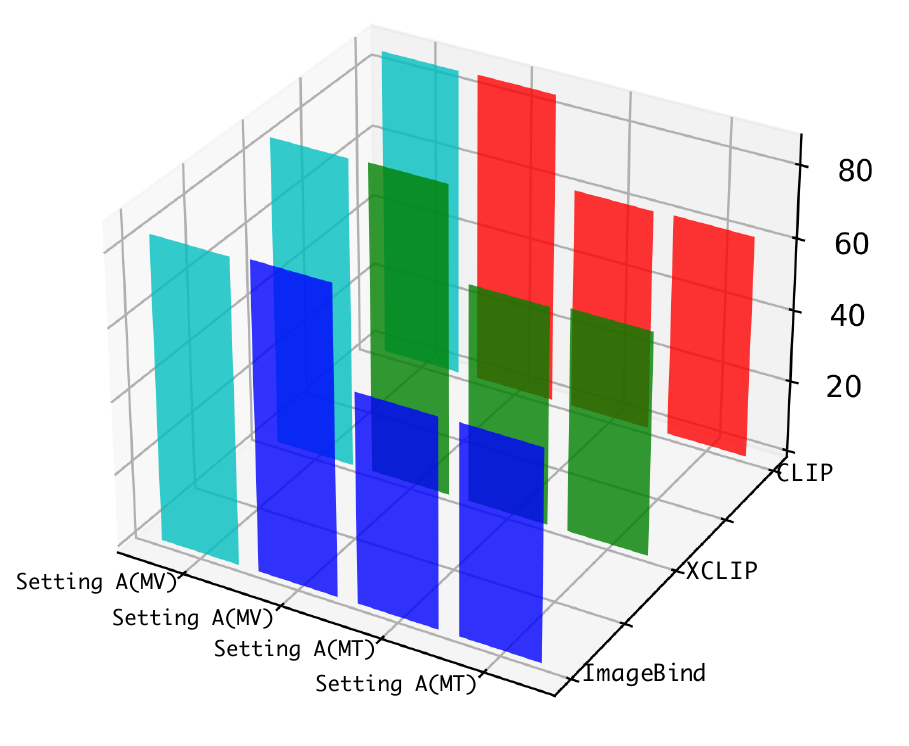}
    \caption{Performance of different multimodal foundational models in scenarios with missing modality on CMU-MOSI dataset.}
    \label{fig:m-clip}
\end{figure}

\begin{figure}
    \centering
    \includegraphics[width=\linewidth]{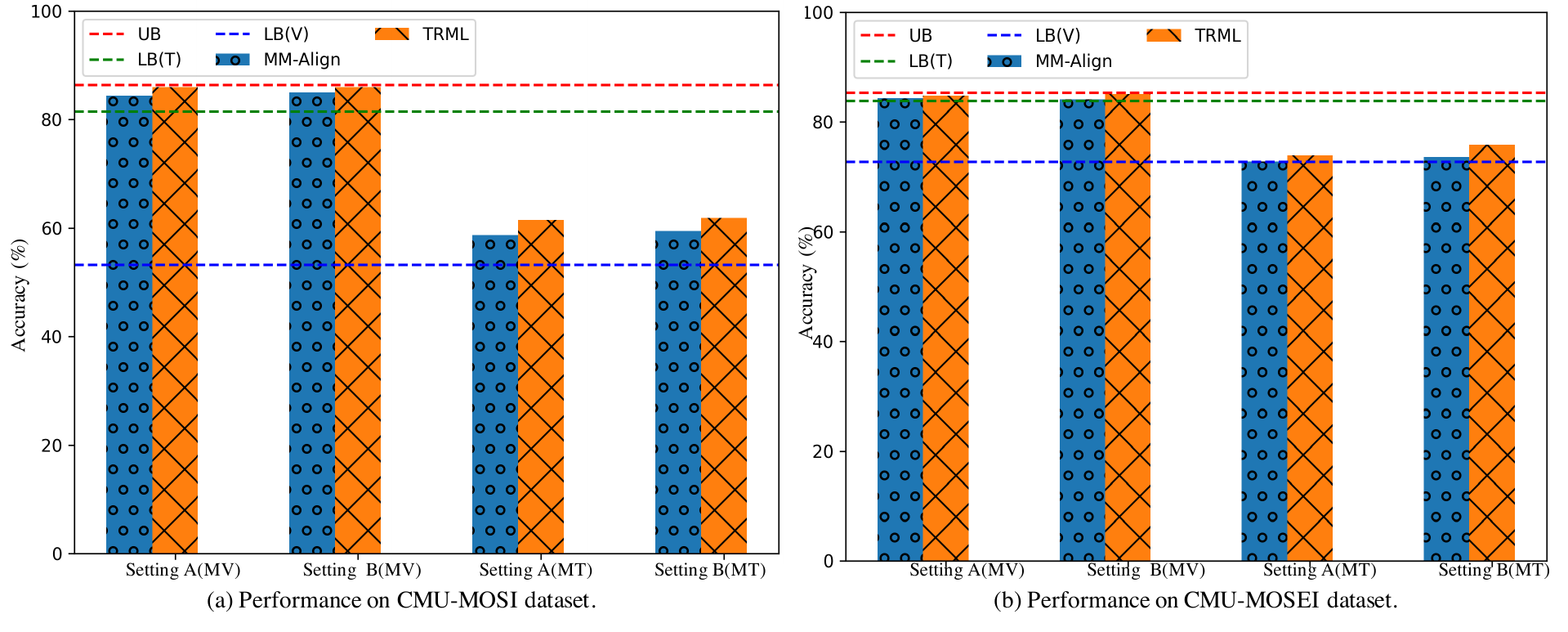}
    \caption{Performance comparisons of different model using the same multimodal foundational model on CMU-MOSI and CMU-MOSEI. The pvalue$<$0.05 of significance test(t-test) in all Setting.}
    \label{fig:mmalign}
\end{figure}

\begin{figure}
    \centering
    \includegraphics[width=\linewidth]{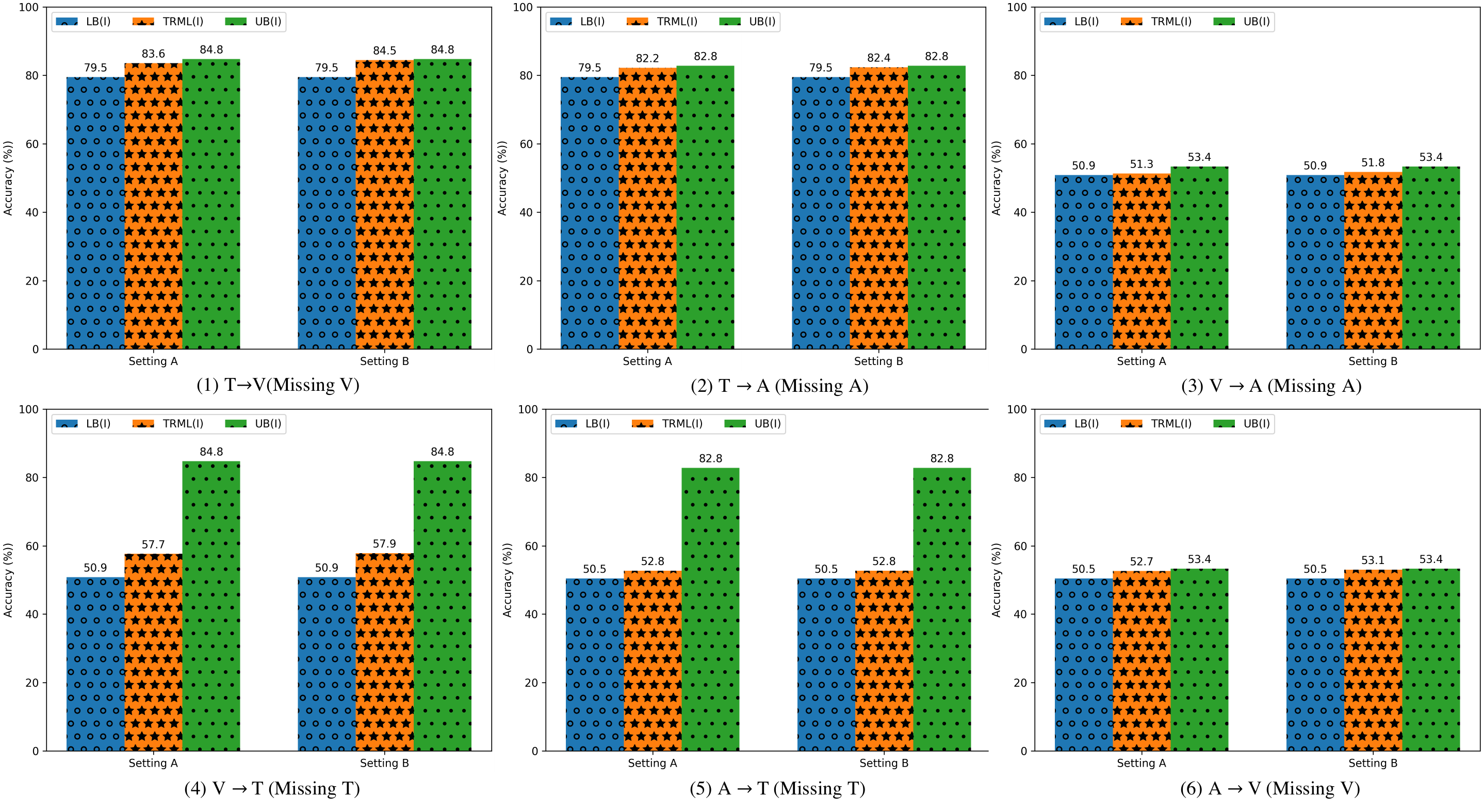}
    \caption{Performance of Imagebind with acoustic modality on CMU-MOSI dataset.}
    \label{fig:imagebind}
\end{figure}

Table \ref{tab:ablation-t} and \ref{tab:ablation-v} display the results of ablation experiments. The experimental results demonstrate the effectiveness of both the unimodal matching loss and semantic matching loss.  Figure \ref{fig:m-clip} illustrates the performance of various multimodal foundational models embedded in TRML. This observation indicates that our proposed framework is flexible, showcasing its ability to enhance the robustness of any clip-based multimodal foundational models. Additionally, we also compared the performance of our model using multimodal foundational models with the current state-of-the-art methods MM-Align. Figure \ref{fig:mmalign} displays the experimental results. These results indicates that our model outperforms MM-Align and is closer to the upper bound compared to MM-Align. We also conducted significance tests (t-test), obtaining p-values all below 0.05 (pvalue$<$0.05).

\begin{table}[]
    \centering
    \tiny
    \begin{tabular}{cccccccccc}
    \toprule
    \multirow{3}{*}{Variants} & \multirow{3}{*}{Dataset} & \multicolumn{4}{c}{V$\rightarrow$T (Missing text modality.)}\\
    \cmidrule(r){3-6}
    & & \multicolumn{2}{c}{Setting A} & \multicolumn{2}{c}{Setting B}  \\
    \cmidrule(r){3-4} \cmidrule(r){5-6}
    & & MAE $\downarrow$ & Acc-2 $\uparrow$ & MAE $\downarrow$ & Acc-2 $\uparrow$  \\
    \midrule
    \textbf{TRML}  & CMU-MOSI  & \textbf{1.410} & \textbf{61.5} & \textbf{1.408} & \textbf{61.9} \\
    $w / o \mathcal{L}^{(t, \bar{t})}$ & CMU-MOSI & 1.438 & 57.4 & 1.441 & 57.7 \\
    $w / o \mathcal{L}^{(v, \bar{v})}$ & CMU-MOSI  & 1.428 & 59.0 & 1.425 & 59.1 \\
    $w / o \mathcal{L}_{s m l}$ & CMU-MOSI  & 1.444 & 56.7 & 1.466 & 56.4 \\
    \midrule
    \textbf{TRML}  & CMU-MOSEI & \textbf{0.856} & \textbf{73.9} & \textbf{0.818} & \textbf{75.8} \\
    $w / o \mathcal{L}^{(t, \bar{t})}$ & CMU-MOSEI  & 0.895 & 73.1 & 0.833 & 74.1 \\
    $w / o \mathcal{L}^{(v, \bar{v})}$ & CMU-MOSEI  & 0.891 & 73.4 & 0.829 & 74.2 \\
    $w / o \mathcal{L}_{s m l}$ & CMU-MOSEI  & 0.897 & 73.0 & 0.890 & 73.4 \\
    \bottomrule
    \end{tabular}
    \caption{Ablation experiments for missing text modality $T$.}
    \label{tab:ablation-v}
\end{table}

\subsection{Semantic Matching Analysis}
To gain insight into how our model aligns cross-modal semantic spaces, we visualize the similarity matrices between different modalities. 
We randomly select text and visual modality pairs from 8 samples in the CMU-MOSI dataset. Figure \ref{fig:similarity} displays the heatmap of the similarity matrices. Figures \ref{fig:similarity}(a)$\sim$(e) respectively demonstrate the similarity between the original modality and the virtual modality. Figures (a)$\sim$(b) indicate a consistent trend in similarity changes between the generated modalities and the original modalities. The results in Figures (c)$\sim$(d) suggest a very high semantic similarity between our generated modalities and the original modalities. Figures (d)$\sim$(e) demonstrate that the virtual modality and the original modality exhibit high cross-modal semantic similarity. In summary, the virtual modalities generated by our model can reconstruct the semantic information of the original modalities across multiple dimensions. This indicates the effectiveness of the missing modality inference and semantic matching learning modules in our framework.

\begin{figure}[h]
  \centering
  \includegraphics[width=\linewidth]{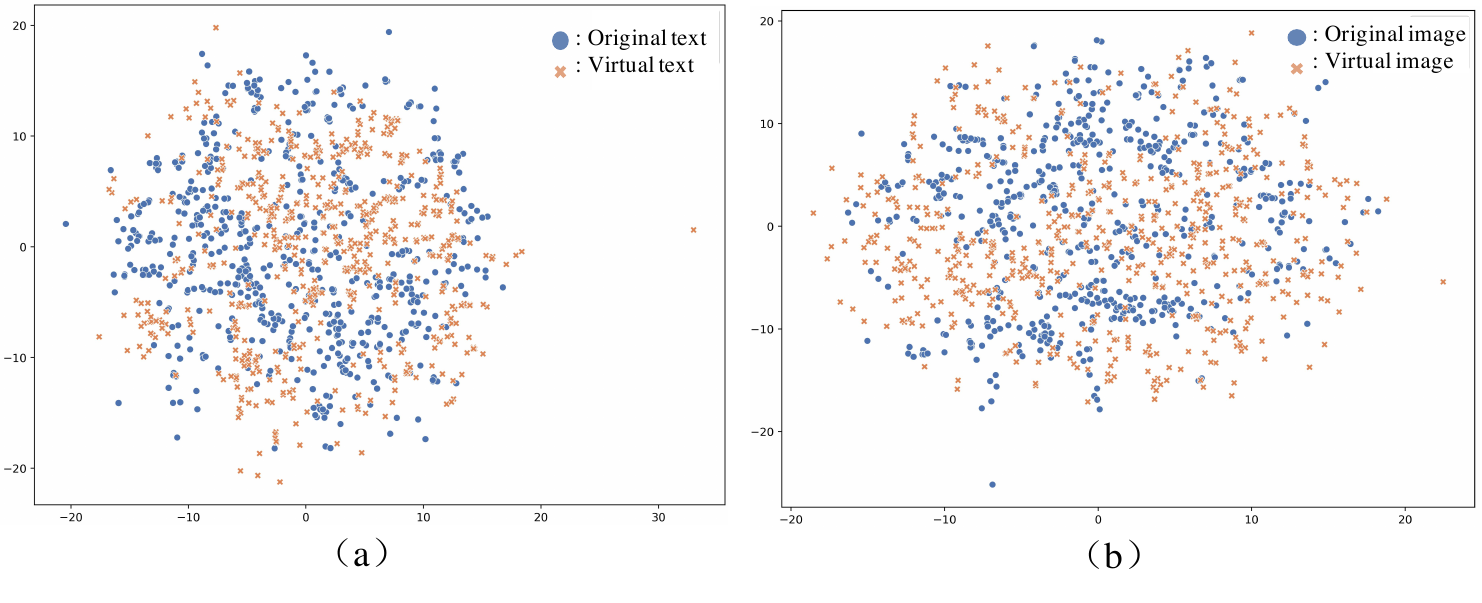}
  \caption{Visulization of the representations of original modality and virtual modality on the CMU-MOSI test dataset.
  } 
  \label{fig:visualization}
\end{figure}
\subsection{Visualization of Virtual and Original modality}
Apart from achieving superior performance, our model can motivate the generalization ability of multimodal foundational models for missing modality inference. The multimodal foundational model is naturally robust due to the technique of contrastive pre-training, and it captures cross-modal latent semantic associations. The key advantage of  our model is to leverage this cross-modal learning ability of multimodal foundational models. To this end, we visualize the generated virtual modality and the original modality. We use t-SNE to map the representations of these modalities into 2D space and visualize them. Each color represents a modality. From Figure \ref{fig:visualization}, we could observe that the representations of the two modalities overlap and are less confined. These results indicate that our proposed framework TRML can well enhance the robustness of multimodal foundational models by bringing original and virtual modality distributions closer.

\subsection{Parameter Analysis}
\begin{figure}
    \centering
    \includegraphics[width=\linewidth]{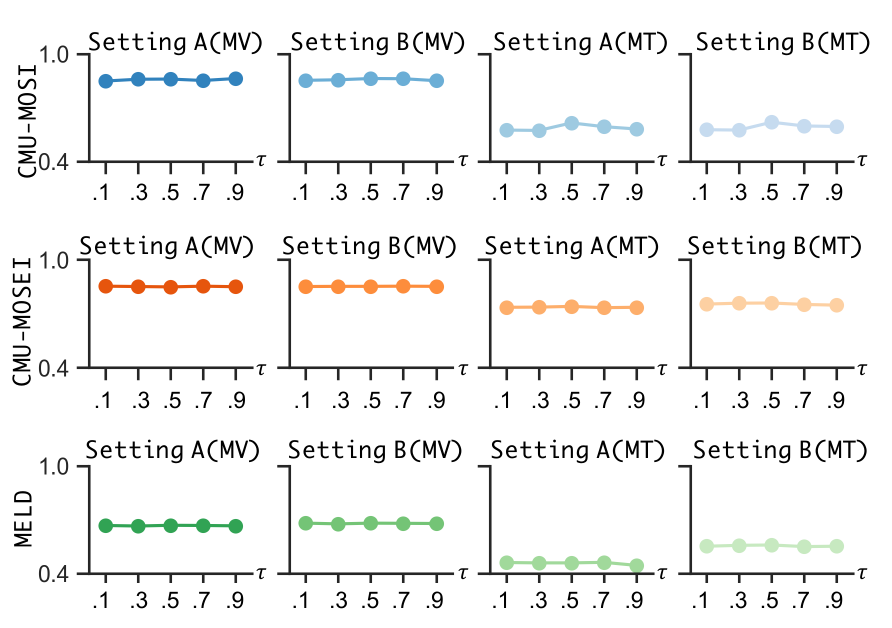}
    \caption{Impact of temperature parameter $\tau$ variation on model performance.}
    \label{fig:para}
\end{figure}

To explore the impact of temperature $\tau$ on model performance, we do experiments by increasing the temperature $\tau$ from 0.1 to 0.9. The results are shown in Fig. \ref{fig:para}. We could draw the following conclusions from the results.
In the absence of the text modality, the model is more sensitive to the temperature parameter $\tau$ compared to the visual modality missing. The reason is that the text modality provides adequate performance when the visual modality is missing, and the poor performance of the vision relative to the text leads to the model being more sensitive to temperature $\tau$ in the Setting B scenario.

\begin{figure}
    \centering
    \includegraphics[width=\linewidth]{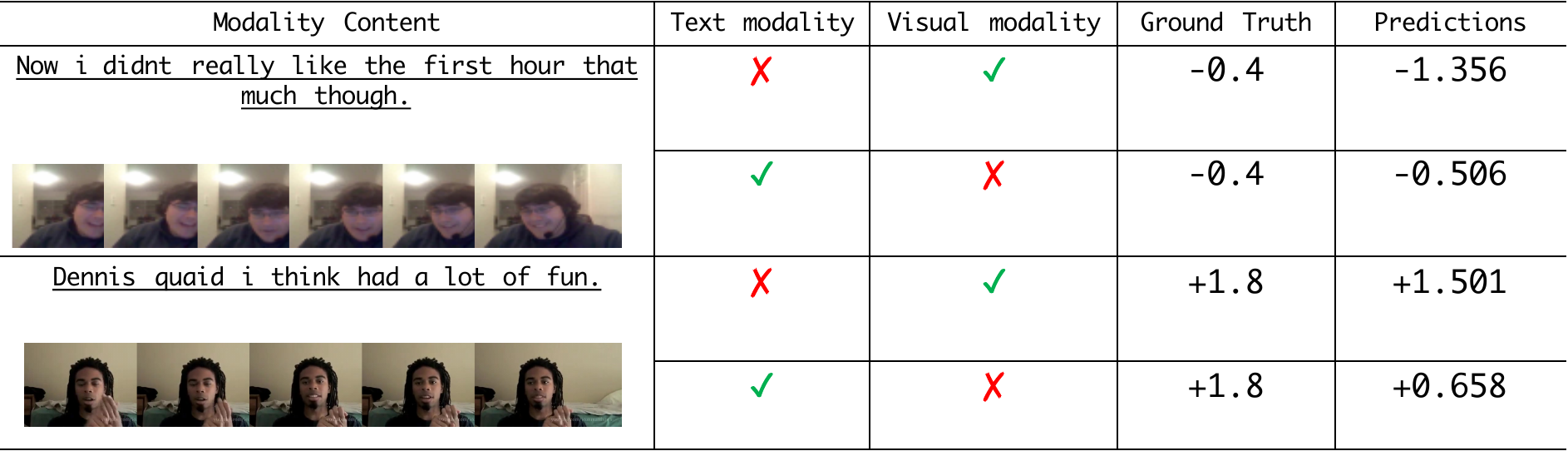}
    \caption{Predictions of TRML with missing different modalities on CMU-MOSI dataset.}
    \label{fig:case}
\end{figure}

\subsection{Case Study}
To qualitatively validate the effectiveness of TRML, we displayed two samples with different sentiment polarities in the CMU-MOSI dataset. 
Based on the sentiment analysis results, we observe that the model can accurately predict sentiment polarity using the generated virtual modality in scenarios with missing modalities. These predictions demonstrates the robustness of our proposed framework. The prediction is more accurate when the visual modality is missing compared to when the text modality is missing. This result can be attributed to the text modality's superior performance compared to the visual modality. The semantic matching module generates virtual text modalities using the original visual modality as prompt. The noise present in the visual modality leads to lower performance in the generated virtual text modality.

\section{Conclusion}
We design a multimodal learning framework that extends CLIP-based multimodal foundational models to handle scenarios with missing modalities. It leverages the multimodal foundational model to generate virtual modalities for traing and test, replacing the missing modalities by aligning cross-modal semantic spaces. Experiments show that our framework can effectively stimulate the robustness of multimodal foundational models with incomplete modality.

\bibliographystyle{named}
\bibliography{ijcai24}

\appendix
\section{Dataset}
\textbf{CMU-MOSI} \cite{zadeh2016mosi} is a multimodal sentiment dataset consisting of 2199 video clips. These video clips are split into a training set of 1284 samples, a validation set of 229 samples, and a test set of 686 samples. The sentiment label of these samples is manually annotated, which is a sentiment score with values ranging from -3 to 3. 3 and -3 indicate strongly positive and strongly negative sentiment polarity, respectively.
\newline
\textbf{CMU-MOSEI} \cite{zadeh2018memory} is also a multimodal sentiment analysis dataset. It is a collection of movie review video clips from Youtube. The sample size is about ten times larger than CMU-MOSI. Similarly, it is labeled with manually annotated sentiment scores, which take values from -3 (i.e., strongly negative) to 3 (i.e., strongly positive).
\newline
\textbf{MELD} \cite{poria2018meld} is a multimodal sentiment analysis and multimodal emotion recognition dataset. These samples are video clips taken from the Friends TV series. It consists of over 1400 dialogues and 13,000 utterances. Different from CMU-MOSI and CMU-MOSEI, its dialogues involve multiple speakers. Each utterance is manually annotated with sentiment polarity and emotion category. The sentiment polarity and sentiment category are corresponding to (anger, disgust, sadness, joy, neutral, surprise, and fear) and (positive, negative, and neutral). respectively.

\section{Implementation Detail}
We set the batch size to 16, 32, and 32 for datasets CMU-MOSI, CMU-MOSEI, and MELD, respectively. For all datasets, the optimizer adopts Adam and the learning rate is set to 5e-5. The Hyper-parameters $\lambda$ and $\alpha$ are setting to 0.1 and 0.5 respectively. Following previous work\cite{tsai2019multimodal,han2022mm}, for the multimodal sentiment analysis task, we adopt binary accuracy (Acc-2) and mean absolute error (MAE) as the metric. For the multimodal emotion recognition task, we adopt the multi-classification accuracy (Acc) as the metric.

\section{Baselines}
\begin{itemize}
    \item  \textbf{Supervised-Unimodality}. In both Settings A and Setting B, there is a non-victim modality, and the performance of the non-victim modality will be taken as the lower bound (\textbf{LB}).
    \item \textbf{Supervised-Bimodality}. We consider the scenario where the multimodal data is not missing and the performance that can be achieved is an upper bound \textbf{UB} by using the complete multimodal data.
    \item \textbf{MFM} \cite{tsai2019learning} employs modality-specific generative factors to capture modality-specific information and generate modality.
    \item \textbf{SMIL} \cite{ma2021smil} employs clustering methods to learn prior information about modalities to generate missing modalities.
    \item \textbf{Modal-Trans} \cite{wang2020transmodality} employ a cyclic sequence-to-sequence
model to capture bidirectional reconstruction information of two modalities.
    \item \textbf{MM-Align} \cite{han2022mm} formulates cross-modal information transfer as an optimal transport problem, which captures and imitates dynamic alignment of cross-modal semantics.
    
\end{itemize}

\end{document}